%% file: main.tex
\documentclass[11pt]{article}

\usepackage{acl}

\usepackage{times}
\usepackage{latexsym}
\usepackage[T1]{fontenc}
\usepackage[utf8]{inputenc}
\usepackage{microtype}

\usepackage{graphicx}
\usepackage{booktabs}
\usepackage{colortbl}
\usepackage{pifont}
\definecolor{bestrow}{gray}{0.88}
\definecolor{lightgray}{gray}{0.93}

\usepackage{tabularx}
\usepackage{array}
\usepackage{multirow}
\usepackage{amsmath,amssymb}
\usepackage{enumitem}
\usepackage{float}
\usepackage{xspace}

\newcommand{\nbf}[1]{{\noindent \textbf{#1}}}
\newcolumntype{L}[1]{>{\raggedright\arraybackslash}p{#1}}

\usepackage[most]{tcolorbox}

\usepackage[capitalize]{cleveref}
\crefname{appendix}{App.}{App.}
\crefname{section}{Sec.}{Secs.}
\crefname{table}{Table}{Tables}
\crefname{figure}{Fig.}{Figs.}
\crefname{algocf}{alg.}{algs.}
\Crefname{algocf}{Algorithm}{Algorithms}

\setlist{nosep,topsep=2pt,partopsep=0pt,itemsep=2pt,parsep=0pt,leftmargin=1.3em}
\newcolumntype{Y}{>{\raggedright\arraybackslash}X}
\newcolumntype{C}{>{\centering\arraybackslash}p{0.11\linewidth}}

\newcommand{\bench}{\texttt{E2A-Bench}\xspace}
\newcommand{\ucr}{\textsc{UCR}\xspace}
\newcommand{\rci}{\textsc{RCI}\xspace}
\newcommand{\eci}{\textsc{ECI}\xspace}
\newcommand{\ndr}{\textsc{NDR}\xspace}
\newcommand{\covdir}{\mathrm{Cov}_{\mathrm{dir}}\xspace}

\newcommand{\buy}{\ensuremath{\mathrm{BUY}}\xspace}
\newcommand{\sell}{\ensuremath{\mathrm{SELL}}\xspace}
\newcommand{\hold}{\ensuremath{\mathrm{HOLD}}\xspace}
\newcommand{\uncertain}{\ensuremath{\mathrm{UNCERTAIN}}\xspace}

\definecolor{bestcov}{HTML}{E3F4E8} 
\definecolor{bestndr}{HTML}{DDEBFF} 
\definecolor{gainrow}{HTML}{EAF4EA}

\title{\bench: Benchmarking Evidence-to-Action Reliability in Financial Chart Reasoning}

\def\authorBlock{
    Xiaoya Wang$^{\textnormal{1,2}}$\footnotemark[1] \qquad
    Yutong Xu$^{\textnormal{1,2}}$\footnotemark[1] \qquad
    Junjie Wang$^{\textnormal{1}}$\footnotemark[1]\footnotemark[2] \\
    $^{\textnormal{1}}$Tsinghua University \qquad
    $^{\textnormal{2}}$Jinan University \\
    {\tt\small xiaoyawang1202@126.com} \quad
    {\tt\small xuyutong1209@gmail.com} \quad
    {\tt\small wangjunjie@sz.tsinghua.edu.cn}
}

\author{\authorBlock}

\begin{document}
\maketitle

{
  \renewcommand{\thefootnote}%
  {\fnsymbol{footnote}}
  \footnotetext[1]{Equal contribution. Xiaoya Wang and Yutong Xu conducted this work as research assistants at Tsinghua University.}
  \footnotetext[2]{Corresponding Author.}
}

\begin{abstract}
Can financial vision--language models (VLMs) turn chart evidence into reliable action recommendations?
Existing hallucination evaluations are mostly claim-centric: they assess whether generated statements are supported, but not whether evidence remains traceable through rationale, confidence, and final action.
We introduce \bench, a $969$-query benchmark for financial chart reasoning, constructed from $323$ HS300 constituents under three input modalities with deterministic OHLCV-derived evidence anchors.
\bench evaluates grounding, reasoning--action consistency, evidence--confidence calibration, and directional coverage through \ucr, \rci, \eci, and \ndr, where \ndr{} measures coverage-aware evidence-to-action reliability rather than realized trading performance.
Evaluating $20$ VLMs reveals three failures hidden by scalar hallucination scores: the lowest-\ucr{} model ranks near the bottom by \ndr{} due to only $6.4\%$ directional coverage; oracle-aided verification reduces unsupported claims but can collapse coverage; and financial fine-tuning amplifies the \buy{}:\sell{} ratio by $4.21$--$4.68\times$ across strict base--fine-tuned pairs.
These results show that financial VLM evaluation should trace the full evidence-to-action chain rather than rely on a single hallucination score.
Code and data: \url{https://github.com/wanng-ide/E2A-Bench}.
\end{abstract}

\section{Introduction}

\begin{figure*}[!t]
\centering
\includegraphics[width=\textwidth]{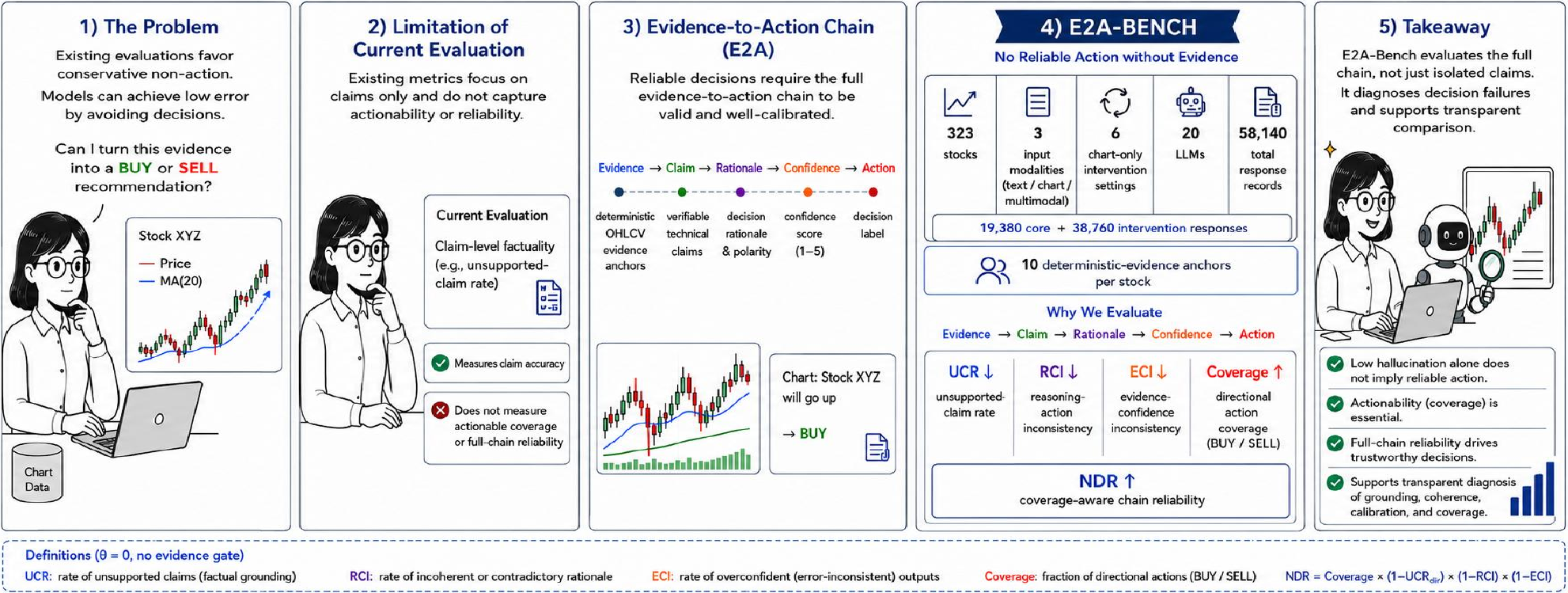}
\caption{Overview of coverage-aware evidence-to-action evaluation in E2A-Bench. 
Existing evaluations are often coverage-blind and chain-blind: they measure whether claims are supported, but not whether models preserve actionable coverage or maintain full-chain reliability. 
\bench addresses both by evaluating evidence-to-action reliability through grounding, coherence, calibration, and net decision reliability.}
\label{fig:chain}
\end{figure*}

In financial chart reasoning, such as A-share technical analysis, vision--language models (VLMs) are expected to identify technical signals such as candlestick patterns, moving-average crossovers, and volume changes~\citep{huang2025chart_understanding_survey,lee2025finllms,ding2024llm_trading_survey}.
They may further issue an action recommendation among \buy{}, \sell{}, \hold{}, and \uncertain{}, together with a confidence estimate.
Reliability in this setting is therefore not reducible to factual correctness at the text level.
A decision-reliable model must ground its recommendation in verifiable chart evidence without systematically avoiding directional decisions when sufficient evidence exists.
Otherwise, a fabricated ``golden cross,'' volume breakout, or trend reversal may become the stated rationale for a \buy{} recommendation, turning a local hallucination into a decision-level risk.

However, existing evaluations do not fully capture this requirement.
Chart hallucination benchmarks are mostly claim-centric, assessing whether local statements are supported by visual or structured evidence, whereas financial multimodal benchmarks are typically task-centric, reporting aggregate scores or final-answer accuracy~\citep{wang2024charxiv,xue2024famma,liu2025visfineval,shu2025finchartbench}.
Both perspectives are useful, but they share an \emph{evaluation-unit mismatch}: they evaluate claim correctness or task outcomes rather than \emph{evidence-backed action}.
This mismatch yields two blind spots: \emph{coverage blindness}, where hallucination scores can favor models that avoid directional recommendations, and \emph{chain blindness}, where evaluations do not verify whether evidence remains consistent from chart-derived facts to claims, rationales, confidence estimates, and final actions~\citep{geifman2017selective,wang2025charthal,halogen}.
Consequently, a model may appear reliable by producing few unsupported claims while offering little actionable value, or by issuing plausible rationales whose actions or confidence estimates are not supported by the underlying evidence.

We frame this issue as a \textbf{coverage--grounding frontier}, as shown in~\cref{fig:chain}: financial VLM reliability should be measured not by unsupported-claim reduction alone, but by whether a model improves evidence grounding while preserving actionable coverage~\citep{geifman2017selective,liu2025uncertainty_calibration_survey}.
A method that lowers hallucination by suppressing \buy or \sell recommendations does not necessarily improve decision reliability.
A meaningful improvement should move models toward the favorable region of this frontier, where they act when evidence is sufficient and ground their actions in verifiable evidence, coherent rationales, and calibrated confidence~\citep{geng2024confidence_calibration_survey,liu2025uncertainty_calibration_survey}.
This framing also clarifies the scope of our evaluation: we do not assess whether models predict market returns, but whether their chart-based recommendations are supported by a verifiable evidence chain.
The protocol is modular: domain-specific evidence anchors, claim patterns, and action schemas can be replaced while preserving the evidence-to-action audit structure.

To operationalize this frontier, we introduce \bench, a $969$-query benchmark over $323$ HS300 constituents and three modalities: text-only, chart-only, and multimodal.
Each query uses a $30$-trading-day OHLCV window and ten deterministic evidence anchors, yielding $3{,}230$ ground-truth facts for claim verification and evidence-strength estimation.
Models produce a technical analysis, rationale, action, and confidence score.
\bench evaluates the resulting evidence-to-action chain with \ucr{} for unsupported claims, \rci{} for reasoning--action inconsistency, \eci{} for overconfidence under weak evidence, and \ndr{} for coverage-aware decision reliability.

We instantiate \bench by evaluating $20$ VLMs spanning proprietary, open-weight, and financially adapted systems.
The experiments reveal three patterns.
First, scalar hallucination scores can misrank decision reliability: Gemma-4-E2B-It has the lowest overall unsupported-claim rate ($39.9\%$), but only $6.4\%$ directional coverage and \ndr{} $=5.1\%$, whereas Qwen2.5-VL-3B-Instruct achieves the highest \ndr{} ($32.1\%$) with $82.5\%$ coverage.
Second, in chart-only intervention analysis, oracle-aided verification can sharply reduce unsupported claims but collapse actionable coverage, yielding negative mean $\Delta\ndr$.
Third, financial fine-tuning redistributes risk: strict base--fine-tuned pairs show inconsistent \ndr{} changes but consistent \buy{}:\sell{} amplification of $4.21$--$4.68\times$.
Together, these results support evaluating financial VLMs along the coverage--grounding frontier rather than relying on scalar hallucination scores.

Our contributions are threefold.
\begin{itemize}[nosep,leftmargin=*]
\item We introduce \bench, a $969$-query benchmark for coverage-aware evidence-to-action reliability in financial chart reasoning.
\item We propose a chain-level protocol for grounding, reasoning--action consistency, evidence--confidence calibration, and directional coverage.
\item We evaluate $20$ VLMs and uncover three failures: coverage-driven rank reversal, verification-induced coverage collapse, and \buy{}-side amplification after financial fine-tuning.
\end{itemize}

\begin{table}[!t]
\centering
\scriptsize
\setlength{\tabcolsep}{2.3pt}
\renewcommand{\arraystretch}{1.10}
\begin{tabular}{
p{1.4cm}
p{1.2cm}
c
p{2.35cm}
c c c}
\toprule
\textbf{Benchmark} 
& \textbf{Scale} 
& \textbf{\#Inp.} 
& \textbf{Main evaluation target} 
& \textbf{Ev.} 
& \textbf{Act.} 
& \textbf{Cov.} \\
\midrule
CharXiv
& $\approx$2K charts
& 1
& Realistic chart QA
& \ding{55} & \ding{55} & \ding{55} \\

ChartHal
& $\approx$2K claims
& 1
& Hallucination detection
& \checkmark & \ding{55} & \ding{55} \\ \midrule

VisFinEval
& $\approx$1K Q\&A
& 1
& Financial chart QA
& \ding{55} & \ding{55} & \ding{55} \\

FinChart-Bench
& $\approx$5K Q\&A
& 1
& Financial chart QA
& \ding{55} & \ding{55} & \ding{55} \\

\rowcolor{blue!7}
\textbf{\bench(ours)}
& \textbf{$\approx$1K Q\&A}
& \textbf{3}
& \textbf{Evidence-to-action reliability}
& \checkmark
& \checkmark
& \checkmark \\
\bottomrule
\end{tabular}
\caption{Comparison with representative chart and financial multimodal benchmarks.
``\#Inp.'' denotes the number of supported input types among text-only, chart-only, and multimodal inputs.
``Ev.'' indicates explicit evidence-grounding evaluation; ``Act.'' indicates decision-oriented action evaluation; and ``Cov.'' indicates coverage-aware scoring.}
\label{tab:benchmark_comparison}
\end{table}

\section{Related Work}

\nbf{Financial chart reasoning with VLMs.}
Recent VLMs have substantially improved chart, document, and structured-image reasoning.
General-purpose model families such as Qwen2.5-VL~\citep{bai2025qwen25vl} and Qwen3-VL~\citep{bai2025qwen3vl} strengthen visual parsing, table/chart understanding, and multimodal reasoning, while chart-oriented methods such as MatCha~\citep{liu2023matcha}, DePlot~\citep{liu2023deplot} and ChartLLaMA~\citep{Dhan2023chartllama} improve chart comprehension through chart derendering, plot-to-table conversion, chart-specific pretraining, or instruction tuning.
Finance-oriented systems such as FinLLaVA~\citep{openfinllms2024finllava} and PyFi~\citep{agenticfinlab2026pyfi} further adapt VLMs to financial tables, charts, and image-based financial reasoning.
These advances make financial chart reasoning increasingly feasible, but they do not establish whether generated recommendations are grounded in verifiable chart evidence, coherent with the stated rationale, calibrated in confidence, and actionable under sufficient evidence.

\nbf{Benchmarking hallucination in financial chart reasoning.}
Existing benchmarks provide important foundations for evaluating chart and financial VLMs.
ChartQA and CharXiv evaluate chart question answering and realistic chart reasoning, ChartMimic evaluates cross-modal chart reasoning, while ChartHal targets fine-grained hallucination in chart understanding~\citep{masry2022chartqa,wang2024charxiv,yang2025chartmimic,wang2025charthal}.
Financial benchmarks such as FAMMA, VisFinEval, and FinChart-Bench extend multimodal evaluation to finance-specific QA, business scenarios, and real-world financial charts~\citep{xue2024famma,liu2025visfineval,shu2025finchartbench}.
Representative chart and financial benchmarks are compared in~\cref{tab:benchmark_comparison}.
However, these evaluations are mostly claim-centric or task-centric: they assess whether local statements are supported, or whether final answers are correct.
Beyond finance, multi-level hallucination diagnostics and evidence-chain evaluation have also been explored in tool use and scientific-document reasoning~\citep{zhang2024toolbehonest,ren2026sinbench}.
\bench is complementary: it shifts the evaluation unit to coverage-aware evidence-to-action reliability, testing whether chart evidence remains traceable through claims, rationales, confidence estimates, and final actions.

\section{\bench: Evidence-to-Action}
\label{sec:benchmark}

To evaluate financial VLMs along the coverage--grounding frontier, we introduce \bench, an evidence-to-action benchmark that tests whether chart-based recommendations are grounded, coherent, calibrated, and actionable.

\subsection{Design Principles}
\label{sec:design_principles}

\bench is guided by three principles.

\nbf{Evidence verifiability.}
Since financial charts do not uniquely determine a correct trading action, we use deterministic OHLCV-derived facts as evidence anchors rather than future returns or human investment judgments.

\nbf{Decision orientation.}
The evaluation target is a recommendation with its rationale and confidence estimate, rather than a descriptive chart summary.

\nbf{Coverage-aware chain integrity.}
Reliable action requires evidence to remain consistent through claims, rationale, confidence, and final action, while avoiding artificial reliability gains from systematic non-action.

\subsection{Task Formulation}
\label{sec:task_formulation}

For each stock $s$ and evaluation date $d_0$, \bench uses the preceding 30 trading days of OHLCV data, where OHLCV denotes open, high, low, close, and volume.
Let $D_{s,d_0}$ denote this market window and let $\mathcal{T}=\{\mathrm{text},\mathrm{chart},\mathrm{multi}\}$ denote the supported input types.
Each input is constructed by a modality-specific formatter:
\begin{equation}
\small
x_{s,d_0}^{(\tau)}
=
\phi_{\tau}(D_{s,d_0}),
\qquad
\tau \in \mathcal{T},
\label{eq:input_form}
\end{equation}
where $\mathrm{text}$ serializes the OHLCV records, $\mathrm{chart}$ renders the candlestick chart with moving averages and volume bars, and $\mathrm{multi}$ provides both.

Given $x_{s,d_0}^{(\tau)}$, the evaluated VLM $M$ produces
\begin{equation}
\small
r_{s,d_0}^{(\tau)}
=
M\!\left(x_{s,d_0}^{(\tau)}\right)
=
(y,\rho,a,c),
\label{eq:model_response}
\end{equation}
where $a\in\{\buy, \sell, \hold, \uncertain\}$ is the action, $y$ is the technical analysis, $\rho$ is the decision rationale, and $c\in\{1,\ldots,5\}$ is the confidence score.
The task does not assign a ground-truth trading action; instead, it exposes the response components needed to evaluate evidence-backed recommendations.

\subsection{Evidence Anchor Construction}
\label{sec:evidence_anchors}

\begin{table}[!t]
\centering
\small
\begin{tabularx}{\columnwidth}{l Y}
\toprule
\textbf{Fact Key} & \textbf{Compact Definition} \\
\midrule
MA alignment & Ordering of MA5, MA10, and MA20 on the last day \\
MA cross & Golden/death cross of MA5 and MA10 in the last 5 days \\
Volume change & Current volume relative to the 20-day mean \\
Price breakout & Current close relative to the preceding 20-trading-day
high--low band, excluding the evaluation day \\
Trend direction & Linear-regression slope of 20-day close prices \\
Volatility & Annualized volatility of recent returns \\
Recent 5-day change & Five-day cumulative price movement \\
Vol--price divergence & Directional relation between recent price and volume changes \\
Price position & Close percentile in the 20-day high--low range \\
Short-term momentum & Three-day price acceleration pattern \\
\bottomrule
\end{tabularx}
\caption{
Compact summary of the ten deterministic OHLCV-derived evidence anchors used in \bench.
}
\label{tab:factkeys}
\end{table}

Financial charts do not provide a unique ground-truth trading action.
We therefore construct reference evidence at the chart-fact level.
For each market window $D_{s,d_0}$, \bench applies a deterministic fact function:
\begin{equation}
\small
F_{s,d_0}=g(D_{s,d_0}),
\label{eq:evidence_anchors}
\end{equation}
where $F_{s,d_0}$ denotes OHLCV-derived technical facts used as evidence anchors.

As summarized in~\cref{tab:factkeys}, the anchors cover common technical-analysis dimensions, including moving averages, price--volume behavior, trend, volatility, range position, and short-term momentum.
Each anchor is computed directly from open, high, low, close, and volume records using fixed rules.
These anchors serve two purposes: they provide reference facts for verifying model-generated technical claims, and they define evidence strength for confidence calibration.
We additionally audit $200$ randomly sampled chart--label pairs and observe no inconsistency between the rendered charts and their OHLCV-derived labels.

Importantly, $F_{s,d_0}$ is not a trading-action label.
A \buy{} or \sell{} recommendation is not evaluated by future return, but by whether the model's stated claims, rationale, and confidence are consistent with verifiable chart evidence.
This design separates evidence-to-action reliability from market prediction while preserving reproducibility.

\begin{figure}[!t]
\centering
\includegraphics[width=\linewidth]{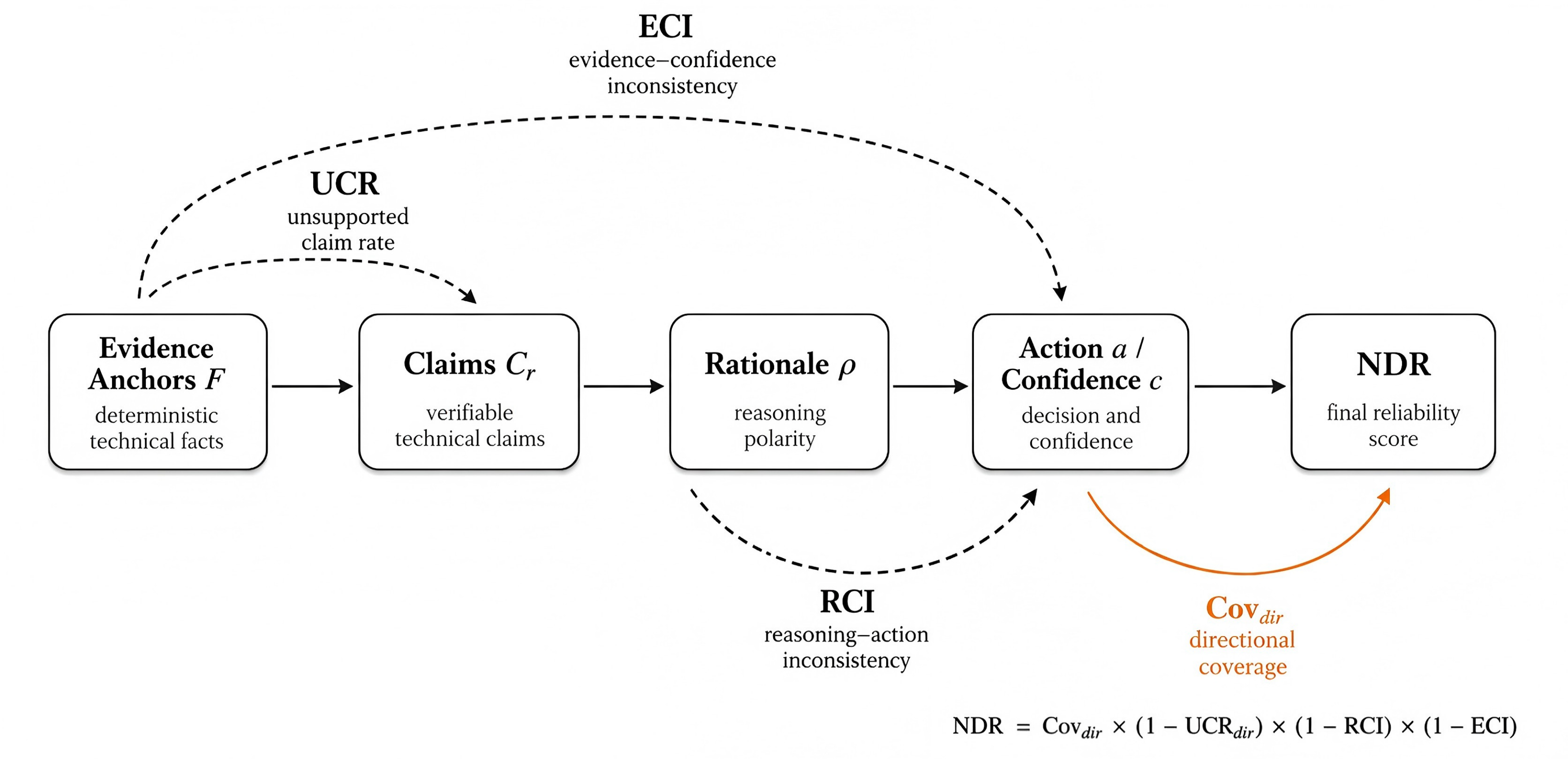}
\caption{
Evidence-to-action scoring protocol in \bench.
A model response is parsed into claims, rationale polarity, action, and confidence; claims are verified against deterministic OHLCV evidence anchors; and the response is scored along grounding, reasoning--action consistency, evidence--confidence calibration, and directional coverage.
}
\label{fig:e2a_chain_protocol}
\end{figure}

\subsection{Evaluation Protocol}
\label{sec:evaluation_protocol}

\paragraph{Overview.}
\bench evaluates each response as an evidence-to-action chain rather than as isolated generated text.
As illustrated in~\cref{fig:e2a_chain_protocol}, the protocol parses a model response, verifies its claims against deterministic evidence anchors, and scores whether evidence remains reliable through the decision process.
Given evidence anchors $F_{s,d_0}$ and a response $r=(y,\rho,a,c)$, the protocol checks four reliability links:
\begin{equation}
\small
\begin{aligned}
F_{s,d_0}\!\rightarrow\!\mathcal{C}_r &: \text{evidence grounding},\\
\rho\!\rightarrow\!a &: \text{reasoning--action consistency},\\
F_{s,d_0}\!\rightarrow\!(a,c) &: \text{evidence--confidence calibration},\\
a\in\mathcal{A}_{\mathrm{dir}} &: \text{directional coverage}.
\end{aligned}
\label{eq:e2a_links}
\end{equation}

\begin{figure*}[!t]
\centering
\includegraphics[width=\textwidth]{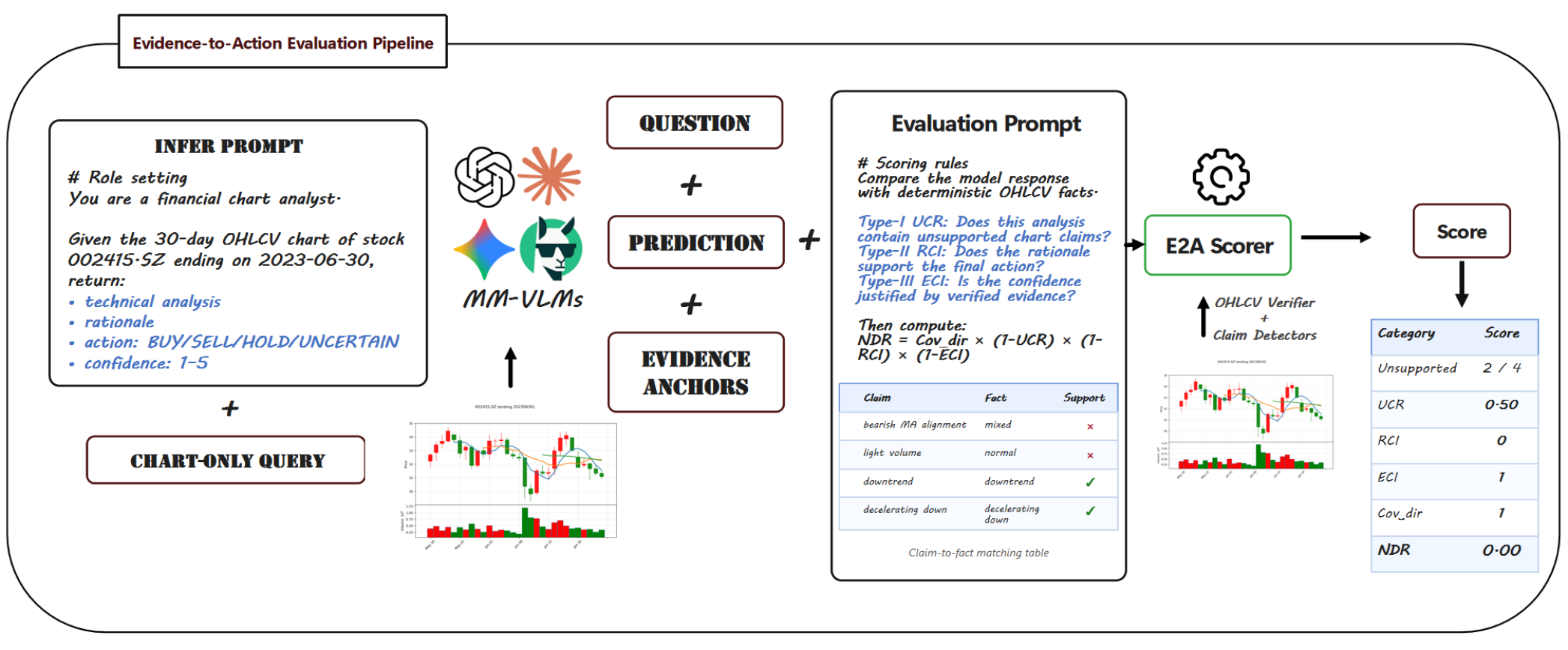}
\caption{
Example of evidence-to-action scoring.
Given a chart-only query, the VLM produces a technical analysis, rationale, action, and confidence score, which are parsed and verified against deterministic OHLCV evidence anchors.
Here, a high-confidence \sell{} recommendation has coherent reasoning but contains two unsupported claims and insufficient evidence for its confidence, yielding $\mathrm{UCR}=0.50$, $\mathrm{RCI}=0$, $\mathrm{ECI}=1$, and $\mathrm{NDR}=0$.
}
\label{fig:e2a_case_pipeline}
\end{figure*}

The first three links define local diagnostic errors, while directional coverage accounts for whether the model preserves actionable outputs.
Together, they instantiate the coverage--grounding frontier: a model is reliable only when it preserves actionable coverage while keeping its decisions grounded, coherent, and calibrated.
\cref{fig:e2a_case_pipeline} illustrates the full scoring process with a chart-only example.

\paragraph{Response parsing.}
Before scoring, each response is normalized into the variables used by the diagnostic protocol:
\begin{equation}
\small
\begin{aligned}
\Pi(r) &= (\mathcal{C}_r,\sigma_r,a,c),\\
\mathcal{C}_r &= \mathrm{Claims}(y,\rho), \qquad
\sigma_r = \sigma(\rho).
\end{aligned}
\label{eq:response_parsing}
\end{equation}
where $\mathcal{C}_r$ is the set of verifiable technical claims extracted from the generated analysis and rationale, with each claim mapped to an evidence anchor.
The variable $\sigma_r\in\{\mathrm{bullish},\mathrm{bearish},\mathrm{neutral},\mathrm{unknown}\}$ denotes the coarse polarity of the rationale.
The parsed action $a$ and confidence score $c$ are used for action-level reliability scoring.
This parsing step exposes response components for evidence-grounded evaluation; it does not judge trading profitability.

\paragraph{Local diagnostic errors.}
Using the parsed tuple $\Pi(r)$ and the matched evidence anchors $F=F_{s,d_0}$, we define three response-level diagnostics for the first three links of the evidence-to-action chain.
Let $\mathbf{1}[\cdot]$ denote the indicator function and let $\mathcal{A}_{\mathrm{dir}}=\{\buy,\sell\}$.

\emph{Unsupported Claim Rate} (\ucr) measures whether the model's verifiable technical claims are supported by the evidence anchors.
Let $\operatorname{Supp}(q,F)$ indicate whether claim $q\in\mathcal{C}_r$ is supported by its corresponding anchor.
We define
\begin{equation}
\small
\mathrm{UCR}(r;F)
=
\frac{
\sum_{q\in \mathcal{C}_r}
\mathbf{1}\!\left[\neg \operatorname{Supp}(q,F)\right]
}{
\max(|\mathcal{C}_r|,1)
}.
\label{eq:ucr}
\end{equation}
When no verifiable claim is extracted, this convention sets $\mathrm{UCR}=0$; such cases are not treated as reliable decisions, but are handled by the coverage term.

\emph{Reasoning--Action Consistency Inconsistency} (\rci) measures whether the final action contradicts the polarity of the stated rationale:
\begin{equation}
\small
\mathrm{RCI}(r)
=
\mathbf{1}\!\left[
(\sigma_r,a)
\in
\{(\mathrm{bullish},\sell),(\mathrm{bearish},\buy)\}
\right].
\label{eq:rci}
\end{equation}

\emph{Evidence--Confidence Inconsistency} (\eci) measures high-confidence directional action under weak evidence.
Let $S(F)$ denote the number of unambiguous directional or extreme signals in the evidence anchors, as specified in~\cref{app:benchmark_spec}.
We define
\begin{equation}
\small
\mathrm{ECI}(r;F)
=
\mathbf{1}\!\left[
S(F)=0
\land a\in\mathcal{A}_{\mathrm{dir}}
\land c\ge 4
\right].
\label{eq:eci}
\end{equation}
Together, \ucr{}, \rci{}, and \eci{} diagnose grounding failure, reasoning--action inconsistency, and evidence--confidence miscalibration, respectively.

\paragraph{Coverage-aware reliability.}
The local diagnostics evaluate the quality of directional decisions, but they do not account for whether the model produces directional actions at all.
For a response $r$ with evidence anchors $F_r$, let $S(F_r)$ denote the number of unambiguous directional or extreme signals.
To support both the main coverage-aware metric and stricter threshold analysis, we define
\begin{equation}
\small
g_{\theta}(F_r)
=
\mathbf{1}\!\left[S(F_r)\ge\theta\right],
\label{eq:evidence_gate}
\end{equation}
where $\theta$ is the minimum strong-signal count required for a directional action to contribute to thresholded coverage.
For a response set $\mathcal{R}$ under a fixed model, modality, and method, the thresholded directional coverage is
\begin{equation}
\small
\mathrm{Cov}_{\mathrm{dir},\theta}
=
\frac{1}{|\mathcal{R}|}
\sum_{r\in\mathcal{R}}
\mathbf{1}\!\left[
a_r\in\mathcal{A}_{\mathrm{dir}}
\land g_{\theta}(F_r)=1
\right].
\label{eq:directional_coverage_theta}
\end{equation}

We compute $\mathrm{UCR}_{\mathrm{dir}}$ by averaging response-level $\mathrm{UCR}$ over directional responses.
By convention, \rci{} and \eci{} also denote directional-conditioned averages unless otherwise specified.
We then define thresholded \emph{Net Decision Reliability} as
\begin{equation}
\small
\begin{aligned}
\mathrm{NDR}_{\theta}
&=
\underbrace{
\mathrm{Cov}_{\mathrm{dir},\theta}
}_{\text{coverage}}
\cdot
\underbrace{
\left(1-\mathrm{UCR}_{\mathrm{dir}}\right)
\left(1-\mathrm{RCI}\right)
\left(1-\mathrm{ECI}\right)
}_{\text{chain reliability}} .
\end{aligned}
\label{eq:ndr}
\end{equation}

The main paper uses $\theta=0$, written as \ndr{} for brevity.
Under this setting, $g_{0}(F_r)=1$ for every response, so $\mathrm{Cov}_{\mathrm{dir},0}$ reduces to ordinary directional coverage, namely the fraction of responses that commit to BUY or SELL.
This setting separates action coverage from chain quality, with grounding, reasoning--action consistency, and evidence--confidence calibration captured by $\mathrm{UCR}_{\mathrm{dir}}$, \rci{}, and \eci{}, respectively.
We analyze $\theta>0$ in \cref{app:threshold} as a stricter robustness analysis that progressively restricts coverage to directional actions supported by stronger chart evidence.
The multiplicative form reflects a necessary-condition view of reliable action: a decision is not reliable if it is absent, unsupported, inconsistent with its rationale, or overconfident under weak evidence.
It should not be interpreted as assuming statistical independence among the factors, nor as measuring realized trading returns.

\paragraph{Reporting convention.}
We report $\mathrm{UCR}_{\mathrm{all}}$ as the average unsupported-claim rate over all responses, which is the closest analogue to a conventional hallucination score.
We also report $\mathrm{UCR}_{\mathrm{dir}}$ to measure grounding quality when the model commits to a directional action.
In contrast, \rci{} and \eci{} are reported on directional responses by default, since their operational meaning concerns action-level consistency and calibration.
Under the main setting $\theta=0$, non-directional outputs such as \hold{} and \uncertain{} reduce directional coverage rather than mechanically lowering consistency or calibration errors.
If no directional response is produced, directional coverage is zero and we set \ndr{} to $0$, while directional-conditioned diagnostics are marked as not applicable.
For $\theta>0$, directional actions that do not satisfy the corresponding evidence threshold are additionally excluded from $\mathrm{Cov}_{\mathrm{dir},\theta}$.
This convention separates evidence-backed action from conservative non-action, which scalar hallucination rates can conflate.

\begin{table}[!t]
\centering
\small
\setlength{\tabcolsep}{4pt}
\renewcommand{\arraystretch}{1.08}
\begin{tabularx}{\columnwidth}{@{}L{0.34\columnwidth} L{0.60\columnwidth}@{}}
\toprule
\textbf{Statistic} & \textbf{Value} \\
\midrule
Stocks & $323$ HS300 constituents \\
Input modalities & text-only, chart-only, multimodal \\
Benchmark queries & $969$ \\
Evidence anchors & $3{,}230$ total facts \\
Input token length, mean $\pm$ std. &
text-only: $1628.8 \pm 61.2$; chart-only: $1027.0 \pm 0.0$; multimodal: $2466.8 \pm 61.2$ \\
\bottomrule
\end{tabularx}
\caption{
Benchmark-level statistics of \bench.
}
\label{tab:benchmark_stats}
\end{table}

\input{tables/oveall-table}

\subsection{Validation and Benchmark Statistics}
\label{sec:validation_statistics}

\nbf{Scorer validation.}
\bench uses deterministic OHLCV-derived evidence anchors, so human annotation is not used to define trading-action ground truth.
We instead use professional annotations from anonymous industry practitioners to validate the automatic scorer.
Annotators judge raw model responses using plain-language questions aligned with \ucr{}, \rci{}, and \eci{}.
The annotations show substantial inter-annotator agreement, and the scorer matches professional consensus labels with high reliability.
In a separate blinded pilot, \ndr{} shows the strongest correlation with holistic human reliability judgments among the compared metrics ($\rho=0.466$, $p=0.009$), while the principal model-level findings remain stable across $1{,}000$ Monte Carlo perturbations of scorer outputs.
Detailed protocols and agreement statistics are reported in~\cref{app:human_validation}.

\nbf{Benchmark statistics.}
\cref{tab:benchmark_stats} reports the reusable benchmark structure: stock universe, input modalities, benchmark queries, evidence anchors, and input-token lengths.
These statistics characterize the scale and input context of \bench; additional distributional details are provided in~\cref{app:benchmark_distribution}.

\section{Experiments and Findings}
\label{sec:experiments}

\subsection{Setup}
\label{sec:setup}

\nbf{Models.}
We evaluate $20$ VLMs across three families: proprietary systems, open-weight models, and financial fine-tunes.
The proprietary set includes Gemini-3.1-Pro/Flash~\citep{google2026gemini31pro,google2025gemini3flash}, GPT-5.5/5.4/5.4-mini~\citep{openai2026gpt54,openai2026gpt55}, and Claude-Sonnet-4.6~\citep{anthropic2026claude46}.
The open-weight set includes Qwen2.5-VL-3B/7B-Instruct~\citep{bai2025qwen25vl}, Qwen3-VL-4B-Thinking~\citep{bai2025qwen3vl}, Qwen3.5-2B/4B/9B/27B~\citep{qwen35blog}, and Gemma-4-E2B/E4B/31B-It~\citep{google2025gemma4}.
The financial set includes Amsi-fin-o1~\citep{Amsi-fin-o1}, FinLLaVA~\citep{openfinllms2024finllava}, and PyFi-QwenVL-3B/7B-COT-47K~\citep{agenticfinlab2026pyfi}.
Details are reported in~\cref{app:experimental_details}.

\subsection{Overall Performance}
\label{sec:overall_performance}

\cref{tab:leaderboard} reports the main trimodal leaderboard across $20$ VLMs.
Overall, reliability is better characterized by the coverage--grounding frontier than by a scalar hallucination score.

\nbf{Low hallucination does not imply reliable action.}
Qwen2.5-VL-3B-Instruct achieves the highest \ndr{} ($32.1\%$) with high directional coverage ($82.5\%$), whereas Gemma-4-E2B-It has the lowest $\mathrm{UCR}_{\mathrm{all}}$ ($39.9\%$) but only $6.4\%$ coverage and \ndr{} $=5.1\%$.
This result shows that low hallucination can reflect conservative non-action rather than evidence-backed decision reliability.

\begin{figure*}[!t]
\centering
\includegraphics[width=0.94\textwidth]{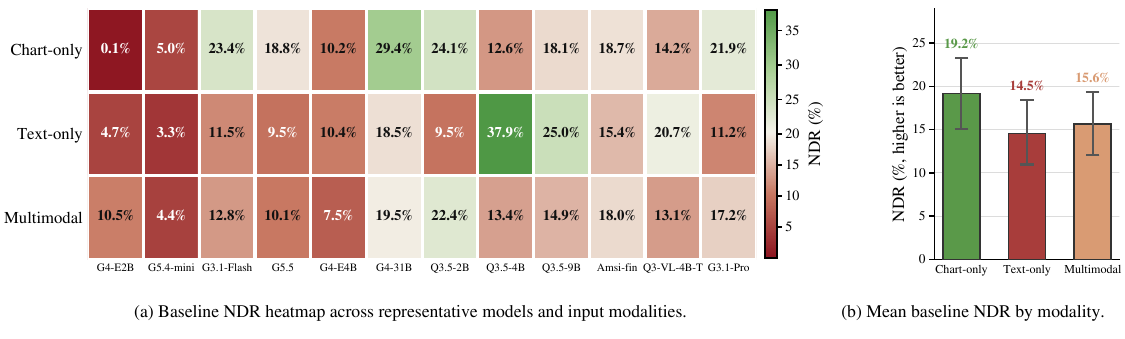}
\caption{
Input modality effects on baseline \ndr.
\emph{(a)} Model-wise \ndr{} across text-only, chart-only, and multimodal inputs for 12 representative models.
\emph{(b)} Mean \ndr{} by modality across all 20 evaluated models, with $95\%$ bootstrap confidence intervals.
Modality effects are model-dependent; chart-only input achieves the highest average baseline \ndr{} across the full model set.
}
\label{fig:ndr_modality}
\vspace{-18pt}
\end{figure*}

\nbf{Model families show distinct reliability profiles.}
Open-weight models achieve the strongest overall reliability, with Qwen2.5-VL-3B/7B ranking first and second by \ndr{}.
Proprietary models are comparatively conservative: the best proprietary model, Gemini-3.1-Pro, reaches only $16.8\%$ \ndr{} despite moderate coverage.
Financially adapted models are competitive but uneven; PyFi-QwenVL-3B reaches $25.3\%$ \ndr{}, while other fine-tuned models are limited by reasoning--action inconsistency, calibration error, or grounding loss.

\nbf{Reliable action requires balancing coverage and chain quality.}
The strongest models do not minimize one diagnostic error in isolation; they preserve actionable coverage while keeping grounding, reasoning--action consistency, and calibration errors controlled.
Financially adapted models illustrate this trade-off: PyFi-QwenVL-3B-COT-47K reaches a competitive \ndr{} ($25.3\%$), but suffers from a high reasoning--action inconsistency rate, whereas Amsi-fin-o1 maintains high coverage ($81.3\%$) but is limited by weaker grounding and calibration.
Among proprietary models, Gemini-3.1-Pro obtains the highest \ndr{} ($16.8\%$), yet remains below the best open-weight and financially adapted systems under our protocol.
Overall, the leaderboard supports the central premise of \bench: decision reliability should be evaluated by movement along the coverage--grounding frontier, not by unsupported-claim rate alone.

\subsection{Diagnostic Analyses}
\label{sec:diagnostic_analyses}

\subsubsection{Input Modality Effects}

\nbf{Input modality affects \ndr{} in a model-dependent manner.}
As shown in~\cref{fig:ndr_modality}, the best input form differs by model: Qwen3.5-4B peaks with text-only input ($37.9\%$), while Gemma-4-31B and Qwen3.5-2B peak with chart-only input ($29.4\%$ and $24.1\%$).
On average, chart-only input gives the strongest baseline \ndr{} ($19.2\%$), outperforming text-only ($14.5\%$) and multimodal input ($15.6\%$).
Therefore, direct chart access can improve evidence-to-action reliability, whereas multimodal input does not consistently yield better decisions, motivating modality-specific reporting.

\begin{figure}[t]
\centering
\includegraphics[width=0.96\linewidth]{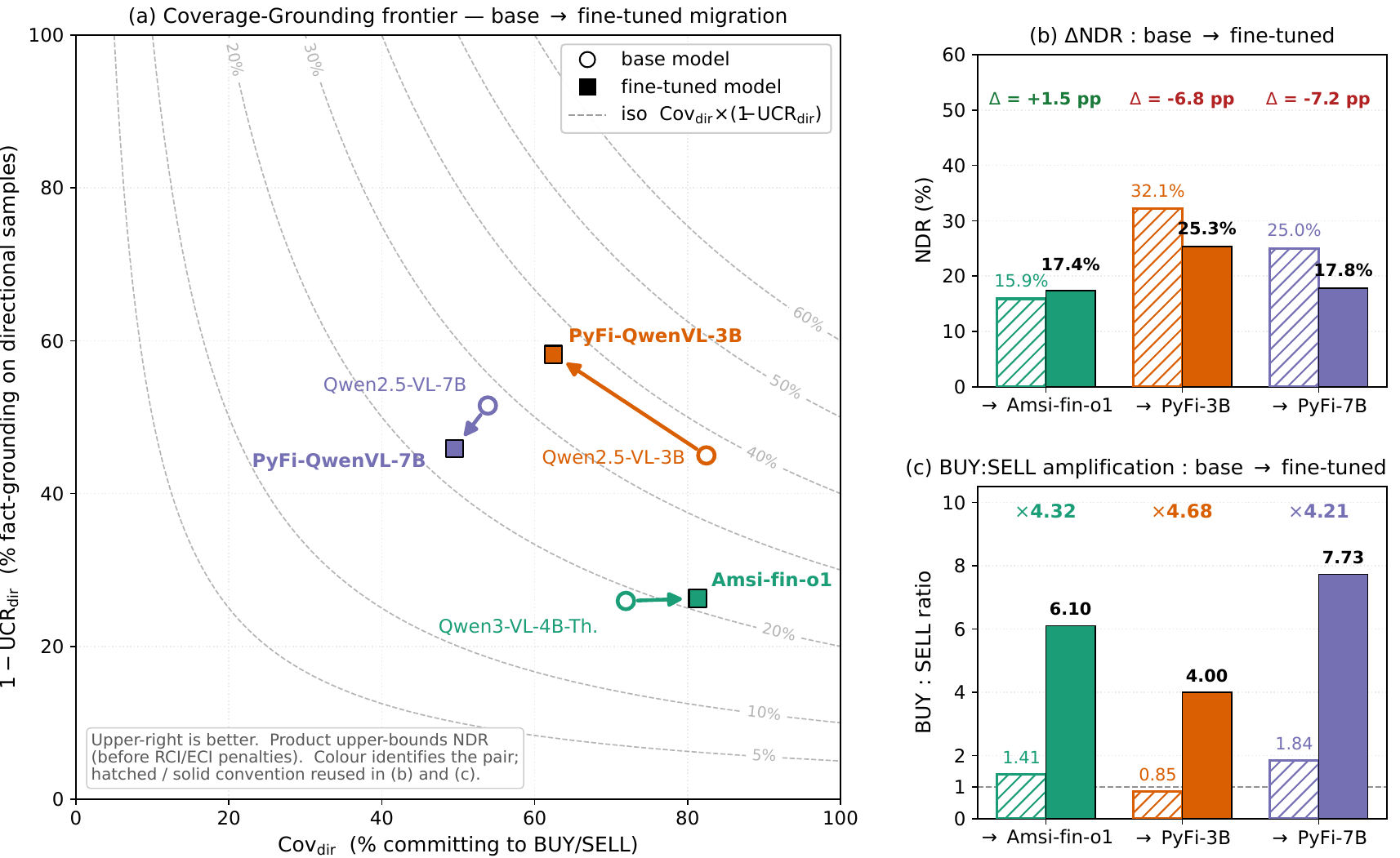}
\caption{
Risk redistribution under financial fine-tuning.
\emph{(a)} Base-to-fine-tuned migration on the coverage--grounding plane, where the dashed curves show iso-$\covdir(1-\mathrm{UCR}_{\mathrm{dir}})$ contours.
\emph{(b)} Fine-tuning produces inconsistent \ndr{} changes across pairs.
\emph{(c)} All pairs show strong \buy{}:\sell{} amplification, indicating a consistent shift toward \buy{} actions despite heterogeneous reliability changes.
}
\label{fig:finetune_frontier}
\vspace{-18pt}
\end{figure}

\begin{figure*}[!t]
\centering
\includegraphics[width=0.8\textwidth]{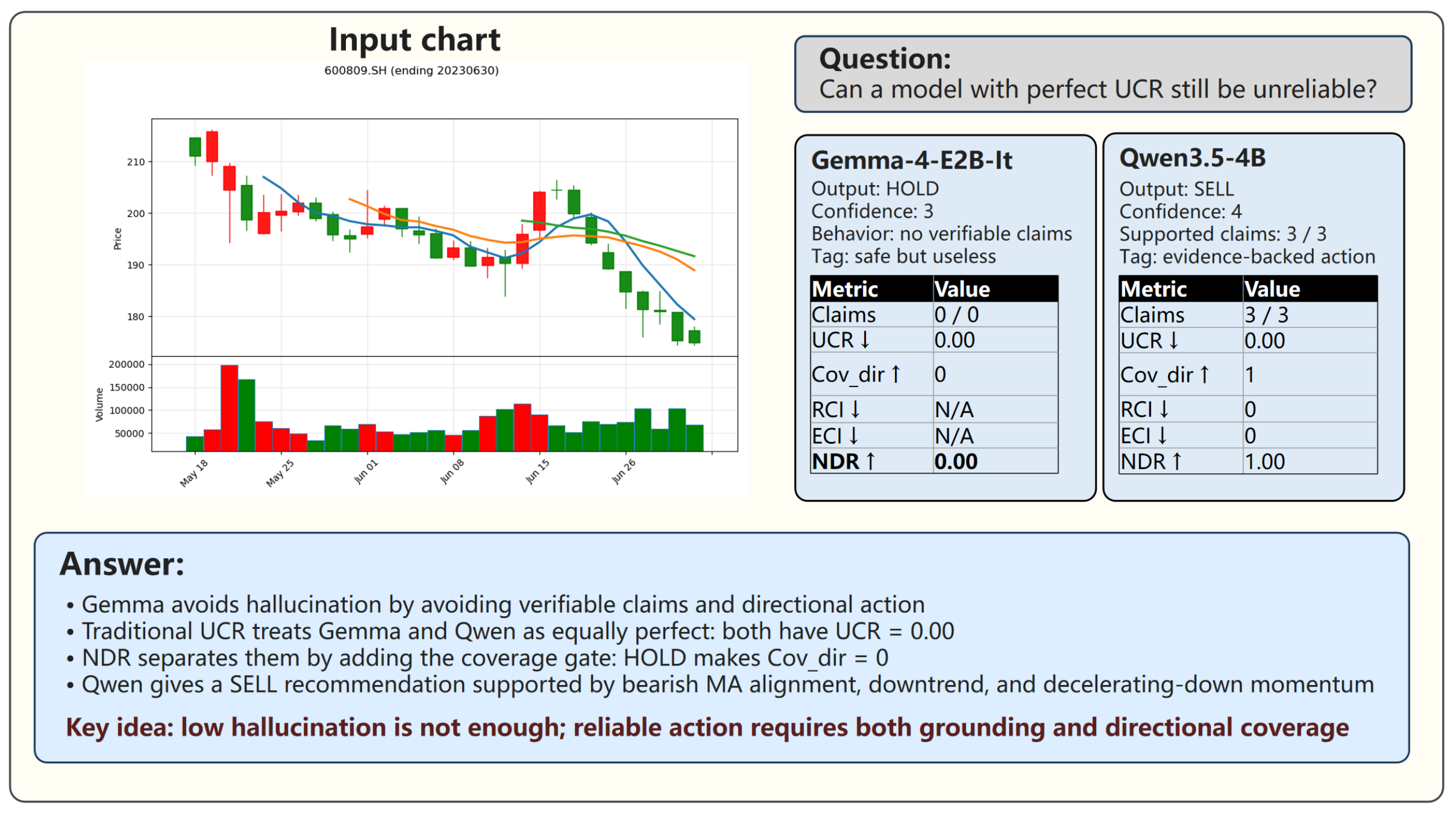}
\caption{
Coverage trap in claim-level hallucination evaluation.
Gemma-4-E2B-It avoids verifiable technical claims and outputs a non-directional \hold{} decision, yielding zero observed \ucr{} but no actionable coverage.
In contrast, Qwen3.5-4B issues a directional \sell{} recommendation whose claims are supported by the evidence anchors.
This case illustrates why claim-level hallucination scores alone can overestimate conservative non-action, while \ndr{} distinguishes non-action from evidence-backed decision reliability.
}
\label{fig:case_coverage_trap}
\vspace{-18pt}
\end{figure*}

\subsubsection{Fine-Tuning Redistributes Risk}
\label{sec:finetuning_analysis}

\cref{fig:finetune_frontier} compares strict base--fine-tuned pairs by their coverage--grounding migration, \ndr{} change, and \buy{}:\sell{} shift.

\nbf{Reliability gains are base-dependent.}
Financial fine-tuning moves different models in different directions on the frontier.
Amsi-fin-o1 improves \ndr{} modestly ($+1.5$ pp), mainly through higher directional coverage with nearly unchanged grounding.
PyFi-QwenVL-3B improves grounding but loses coverage, while PyFi-QwenVL-7B moves downward and leftward; both decrease \ndr{} ($-6.8$ pp and $-7.2$ pp).
Financial fine-tuning redistributes rather than uniformly reduces risk across coverage, grounding, coherence, and calibration.

\nbf{\buy{}-side amplification is consistent.}
Despite heterogeneous \ndr{} changes, all three fine-tuned models shift strongly toward \buy{} actions.
The \buy{}:\sell{} ratio increases from $1.41$ to $6.10$, from $0.85$ to $4.00$, and from $1.84$ to $7.73$, corresponding to $4.21$--$4.68\times$ amplification.
Across the three pairs, $455$ of $2{,}907$ paired responses shift from \hold{}/\uncertain{}/unparseable outputs to \buy{}, while matched evidence claims decrease by approximately $34$--$40\%$.
Because the fine-tuning corpora and objectives are not fully public, we treat these patterns as descriptive evidence of the output shift rather than as a causal explanation.
Together, these results suggest that financial adaptation should be audited not only for aggregate reliability, but also for systematic shifts in action distributions.

\begin{figure}[t]
\centering
\includegraphics[width=\linewidth]{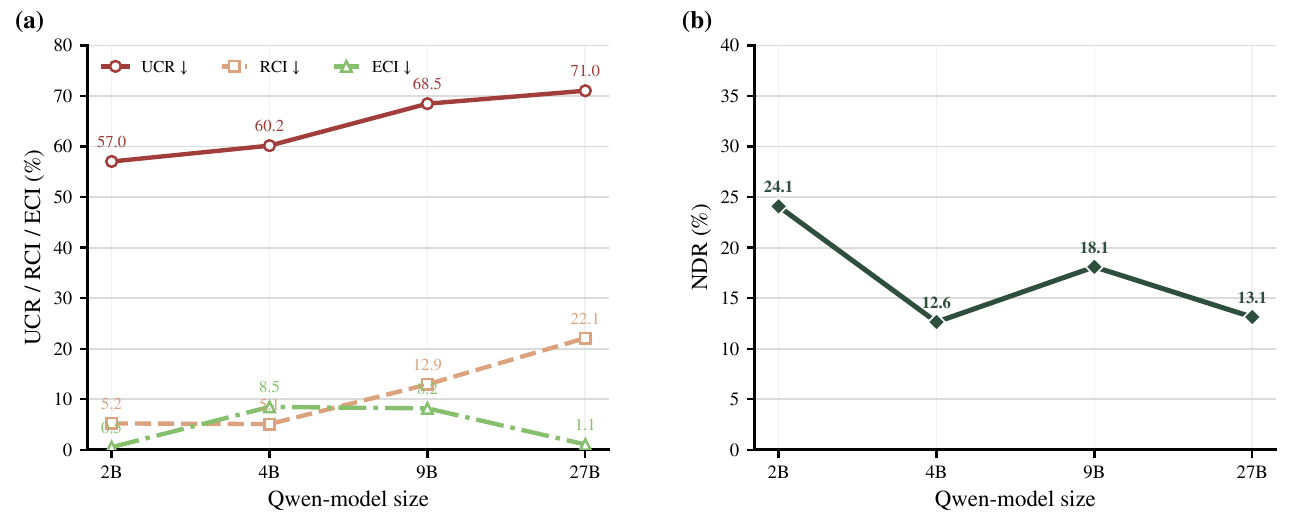}
\caption{
Scaling behavior within Qwen3.5 under chart-only inputs.
\emph{(a)} Local diagnostics (\ucr, \rci, \eci) show non-monotonic and partially conflicting trends across model sizes.
\emph{(b)} \ndr{} also varies non-monotonically, peaking at 2B.
}
\label{fig:qwen_scaling}
\vspace{-18pt}
\end{figure}

\subsubsection{Scaling Is Not Monotonic}
\label{sec:qwen_scaling_analysis}

\nbf{Larger Qwen3.5 models do not consistently improve chain-level reliability.}
We analyze chart-only inputs, a common setting for financial chart reasoning, while holding the model family fixed.
As shown in~\cref{fig:qwen_scaling}(a), scaling from 2B to 27B does not monotonically improve the local diagnostics.
\ucr{} increases from $57.0$ to $71.0$, suggesting weaker factual grounding at larger scales, while \rci{} and \eci{} follow non-monotonic and partially conflicting trends.
Thus, larger models are not uniformly better at preserving grounding, reasoning--action consistency, and confidence calibration.

\nbf{\ndr{} follows a separate scaling pattern.}
\cref{fig:qwen_scaling}(b) shows that \ndr{} peaks at 2B ($24.1$), drops at 4B ($12.6$), partially recovers at 9B ($18.1$), and declines again at 27B ($13.1$).
This trajectory differs from any single local diagnostic and indicates that net decision reliability is shaped by the interaction between coverage and chain quality.
The result reinforces a central point of \bench: parameter scale alone is not a reliable proxy for evidence-to-action reliability.

\subsubsection{Qualitative Illustration}
\label{sec:qualitative_illustration}

\cref{fig:case_coverage_trap} illustrates the coverage trap behind the rank reversal.
Gemma-4-E2B-It obtains zero observed \ucr{} by avoiding verifiable claims and issuing \hold{}, whereas Qwen3.5-4B gives an evidence-backed \sell{} recommendation.
This contrast shows why \ndr{} must account for directional coverage rather than hallucination avoidance alone.

\section{Conclusion}

We introduced \bench, a coverage-aware evidence-to-action benchmark for financial VLM reliability.
By tracing chart evidence through claims, rationale, confidence, and action, \bench reveals failures hidden by scalar hallucination scores.
Our results show that low hallucination can reflect non-action, oracle-aided verification can reduce unsupported claims while collapsing actionable coverage, and financial fine-tuning can amplify \buy{}-side bias.
Importantly, \ndr{} measures evidence-to-action reliability rather than market forecasting or realized trading performance.
Together, these findings suggest that financial VLM evaluation should account for the coverage--grounding frontier rather than rely on hallucination reduction alone.

\section*{Limitations}

\bench is built on HS300 constituents at a fixed evaluation date, so it does not yet cover different market regimes, asset classes, or cross-market trading conventions.
This controlled snapshot improves comparability across models, but future versions should extend the benchmark to bull, bear, and sideways regimes, as well as to broader equity markets and other financial instruments.
Such extensions would test whether evidence-to-action reliability remains stable under changing volatility, liquidity, and macro-market conditions.

\bench relies on deterministic OHLCV-derived evidence anchors and pattern-based claim detectors to ensure reproducible and auditable evaluation.
This design avoids subjective trading-action labels, but it may miss rare paraphrases, nuanced technical claims, or domain-specific reasoning patterns not covered by the current detector.
Future work can improve coverage by expanding multilingual claim patterns, incorporating more robust semantic parsers, and adding professional adjudication for ambiguous cases, while preserving the core principle of grounding evaluation in verifiable evidence.

\section*{Ethical Considerations}

\bench is intended as a research evaluation resource for studying evidence-to-action reliability and should not be interpreted as financial or investment advice.
A high \ndr{} score indicates stronger reliability under our evaluation protocol, but does not imply profitability, investment quality, or readiness for real-world deployment.
Model rankings on \bench should therefore not be used as the sole basis for selecting or deploying systems in financial applications.
Any downstream use of the benchmark or released artifacts should follow applicable financial regulations and the licensing and redistribution requirements of the underlying market-data providers.

\section*{Acknowledgments}

This work was partly supported by the Key Research and Development and Achievement Transformation Program of Inner Mongolia Autonomous Region (Grant No. 2026YFDZ0131).

\bibliography{custom}

\appendix

\section*{Appendix}

\begin{table}[t]
\centering
\small
\begin{tabularx}{\columnwidth}{l Y}
\toprule
\textbf{Signal Type} & \textbf{Qualifying Labels} \\
\midrule
Price breakout & \texttt{breakout\_high}, \texttt{breakout\_low} \\
Volume anomaly & \texttt{heavy\_volume}, \texttt{light\_volume} \\
Strong trend & \texttt{uptrend} or \texttt{downtrend} with $|\mathrm{slope}|>0.30\%$/day \\
Vol--price divergence & \texttt{bullish\_divergence}, \texttt{bearish\_divergence} \\
Extreme price position & \texttt{high\_zone}, \texttt{low\_zone} \\
Momentum acceleration & \texttt{accelerating\_up}, \texttt{accelerating\_down} \\
\bottomrule
\end{tabularx}
\caption{
Evidence-anchor labels counted as strong signals for $S(F)$.
A sample has weak evidence when $S(F)=0$; \eci{} fires when such a sample receives a high-confidence directional action.
}
\label{tab:eci_signals}
\end{table}

\section{Detailed Benchmark Specification}
\label{app:benchmark_spec}

This appendix expands the compact evidence-anchor summary in \cref{tab:factkeys}.
We specify the label spaces, deterministic computation rules, and strong-signal indicators used for evidence-confidence evaluation.
All anchors are computed from the 30-trading-day OHLCV window and serve as evidence labels rather than trading-action labels.

\subsection{Execution Records and Re-run Consistency}
\label{app:storage_consistency}

The core benchmark contains $969$ unique stock--modality queries.
Additional archived response records may arise from chart-only intervention analyses or paired baseline executions.
These records are used for experimental comparison and consistency checking, but are not counted as additional benchmark queries.

For consistency checking, we compare aggregate metrics with and without paired chart-only baseline executions.
Deduplication changes \ucr{}, \rci{}, and \eci{} by at most $0.4$ percentage points and changes \ndr{} by at most $0.005$; model rankings remain unchanged.
Thus, the storage convention does not affect the substantive conclusions.

\subsection{Strong-Signal Rules for Evidence--Confidence Evaluation}
\label{app:eci_signals}

\eci{} penalizes high-confidence directional actions when the chart evidence is weak.
To estimate evidence strength, \bench computes a strong-signal count $S(F)$ from the evidence anchors.
As specified in~\cref{tab:eci_signals}, only directional or extreme labels are counted as strong signals.
Descriptive anchors may still be used for claim verification, but do not by themselves justify high-confidence directional action.

\paragraph{Excluded anchors.}
Not all evidence anchors in~\cref{tab:factkeys_full} contribute to $S(F)$.
MA alignment, volatility, and recent 5-day change are excluded from the strong-signal set because they are descriptive or slowly varying signals rather than standalone directional triggers.
MA alignment may overlap with trend direction, volatility is not directional by itself, and recent 5-day change is treated as contextual evidence.
All three anchors remain available for claim verification under \ucr{}.

\begin{table}[!t]
\centering
\small
\begin{tabular}{lr}
\toprule
\textbf{Quantity} & \textbf{Value} \\
\midrule
Stocks & $323$ \\
Input modalities & $3$ \\
Benchmark queries & $969$ \\
Evaluated models & $20$ \\
Recorded response cells & $19{,}380$ \\
\midrule
Max claim judgments / response & $16$ \\
Response-level judgments / response & $3$ \\
Max scoring slots / response & $19$ \\
Scoring-slot capacity & $368{,}220$ \\
Realized scoring judgments & $112{,}369$ \\
\bottomrule
\end{tabular}
\caption{
Core trimodal benchmark and scoring inventory.
The response-level judgments comprise reasoning--action consistency,
evidence--confidence consistency, and directional coverage.
Chart-only intervention settings are treated as experimental conditions
and are not counted as additional core benchmark queries.
}
\label{tab:trimodal_inventory}
\end{table}

\begin{figure}[!t]
\centering
\includegraphics[width=\linewidth]{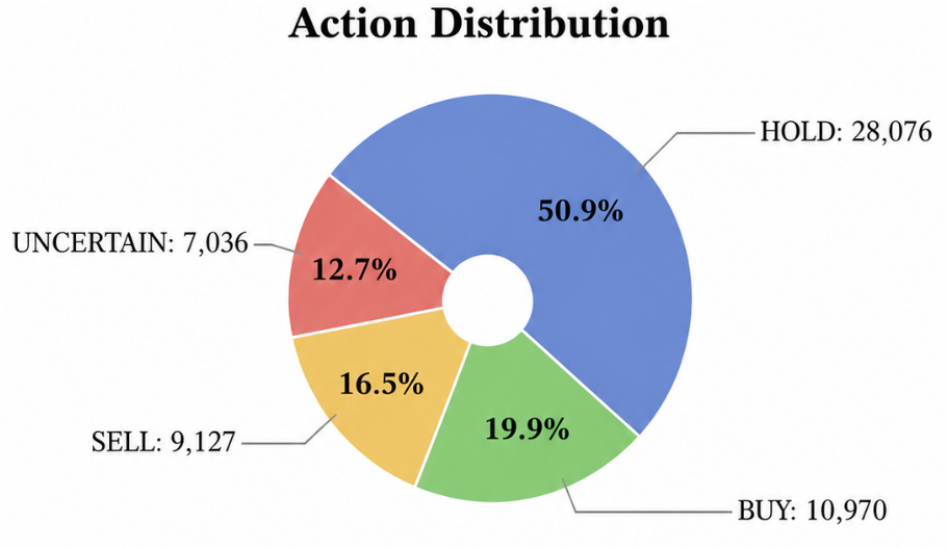}
\caption{
Final-action distribution over 55,209 successfully parsed actions.
\hold and \uncertain account for 63.6\% of parsed actions, motivating coverage-aware evaluation beyond claim-level hallucination.
}
\label{fig:app_benchmark_distributions}
\end{figure}

\input{tables/full-facts}

\subsection{Benchmark Inventory and Output Distribution}
\label{app:benchmark_distribution}

\cref{tab:trimodal_inventory} summarizes the core trimodal inventory of \bench.
The core benchmark contains $323$ stocks under three input modalities, yielding $969$ benchmark queries and $19{,}380$ completed response cells across $20$ models.
Each response exposes up to $19$ scoring slots: at most $16$ claim-grounding judgments, plus three response-level judgments for reasoning--action consistency, evidence--confidence consistency, and directional coverage.

\nbf{Claim-extraction coverage.}
Across all $19{,}380$ responses, $89.2\%$ contain at least one anchor-matched claim, with $2.80$ matched claims per response on average (median $2$); the remaining $10.8\%$ contain no matched claim.
Expressions outside the current claim ontology are left unmatched rather than labeled as unsupported.
\cref{fig:app_benchmark_distributions} shows the final-action distribution over successfully parsed actions.
Non-directional responses dominate: \hold{} and \uncertain{} account for $63.6\%$ of parsed actions.
This motivates coverage-aware scoring, since low unsupported-claim rates can coincide with limited directional action.

\subsection{Full Specification of Evidence Anchors}
\label{app:facts}

\cref{tab:factkeys_full} lists the ten OHLCV-derived evidence anchors used by \bench for claim verification and evidence-strength estimation.

\subsection{Human Validation of the Automatic Scorer}
\label{app:human_validation}

\nbf{Annotation protocol.}
We validate the automatic scorer using professional annotations from four anonymous industry practitioners with finance-related experience.
The annotators participated under written agreements and were informed of the research purpose, intended data usage, annotation procedure, and confidentiality requirements.
Human annotation is used only to validate the response-level scorer; it is not used to define ground-truth trading actions.

Annotators were shown the rendered candlestick chart and the raw model response.
They were not shown model identities, test-time methods, automatic labels, or metric names.
Each response was judged using three plain-language questions: whether the analysis contains unsupported chart claims, whether the reasoning contradicts the final recommendation, and whether the expressed confidence is appropriate given the chart evidence.
These questions correspond to \ucr, \rci, and \eci, respectively.

\begin{table}[!t]
\centering
\small
\begin{tabular}{lcccc}
\toprule
\textbf{Diagnostic} & \textbf{AC1} & \textbf{Precision} & \textbf{Recall} & \textbf{F1} \\
\midrule
\ucr{} & 0.71 & 91.3\% & 88.7\% & 90.0\% \\
\rci{} & 0.74 & 87.6\% & 84.2\% & 85.9\% \\
\eci{} & 0.82 & 93.1\% & 90.5\% & 91.8\% \\
\bottomrule
\end{tabular}
\caption{
Human validation of the automatic scorer.
AC1 measures inter-annotator agreement among four anonymous industry annotators; precision, recall, and F1 are computed against human-consensus labels.
}
\label{tab:validation}
\end{table}

\nbf{Agreement and scorer performance.}
We compute inter-annotator agreement on the overlapping subset using Gwet's AC1.
We use Gwet's AC1 rather than Cohen's $\kappa$ because some error types have low positive-class prevalence, where $\kappa$ can be sensitive to marginal distributions~\citep{wongpakaran2013ac1}.
The four anonymous industry annotators show substantial agreement across all three diagnostics, with AC1 scores of $0.71$ for \ucr{}, $0.74$ for \rci{}, and $0.82$ for \eci{}.
We then evaluate the automatic scorer against human-consensus labels.
As shown in~\cref{tab:validation}, all three diagnostics exceed $85\%$ F1, indicating that the scorer is a reliable proxy for large-scale evidence-to-action evaluation.

The scorer performs strongest on \eci{}, where the signal-count rule is comparatively deterministic.
\rci{} has the lowest recall, reflecting cases with implicit concessive reasoning or weak polarity cues.

\nbf{Holistic human alignment.}
To complement component-level scorer validation, we conduct a blinded pilot on $30$ stratified responses spanning representative reliability profiles.
Two annotators independently judge how reliably each response transforms verifiable chart evidence into its final recommendation, without access to model identity, test-time method, automatic labels, or metric names.
Their ratings show significant agreement ($\rho=0.569$, $p=0.001$), and their mean rating is used as the holistic human judgment.
Among the compared metrics, \ndr{} shows the strongest correlation with this judgment ($\rho=0.466$, $p=0.009$), slightly above directional coverage ($\rho=0.453$) and the coverage--grounding score ($\rho=0.451$).

\nbf{Robustness to scorer uncertainty.}
We further estimate empirical scorer-error patterns from an independently labeled stratified audit and use them to perturb scorer outputs over all $19{,}380$ trimodal responses.
Across $1{,}000$ Monte Carlo trials, Qwen2.5-VL-3B-Instruct remains the highest-\ndr{} model, Gemma-4-E2B-It remains among the bottom three by \ndr{}, and the lowest-\ucr{} model is never the highest-\ndr{} model.
These results indicate that the principal model-level conclusions are robust to scorer uncertainty.

\subsection{Prompt Template and Language Setting}
\label{app:prompt_template}

The raw prompts and model responses in \bench are in Simplified Chinese, matching the A-share market setting and the HS300 stock universe.
For readability, paper examples are translated into English, while all metric computation is performed on the original model outputs.
The framework is language-flexible and can be adapted by replacing the prompt template, response parser, and claim-pattern detectors.
\cref{fig:base_prompt} shows an English translation of the base inference prompt used in the standard financial chart reasoning task.

\begin{figure}[!t]
\centering
\begin{tcolorbox}[
    width=0.96\columnwidth,
    title={Base inference prompt},
    colback=gray!3,
    colframe=black!35,
    fonttitle=\bfseries
]
\small
You are a professional A-share technical analyst.
Please perform technical analysis based on the provided stock data and give a trading recommendation.

\medskip
\textbf{Important:} Only describe information that can be directly observed from the chart or data.
Do not list signals that cannot be verified.

\medskip
Please strictly follow the output format below:

\medskip
\textbf{[Technical Analysis]}\\
Use 3--5 sentences to describe the recent technical trend of the stock.
Only include observations that can be directly read from the chart or data.
Do not use a fixed template or mechanically enumerate signals.

\medskip
\textbf{[Rationale]}\\
Based on the technical analysis above, provide the reasoning behind your overall judgment.

\medskip
\textbf{[Trading Recommendation]}\\
Action: BUY / SELL / HOLD / UNCERTAIN (choose exactly one)\\
Confidence: 1--5 (1 = very uncertain, 5 = very certain)

\medskip
Please answer in Simplified Chinese.
\end{tcolorbox}
\caption{
English translation of the base inference prompt.
The original prompt is written in Simplified Chinese and used for model inference; examples in the paper are translated for readability.
}
\label{fig:base_prompt}
\end{figure}

\subsection{License}
\label{app:license}

We will release the code, evaluation scripts, more prompt templates, processed benchmark metadata, and rendered chart images upon acceptance.
The released artifacts will be distributed under the MIT License.
The chart images in \bench are rendered from publicly available OHLCV market data and do not contain personal or user-generated information.
Raw market data redistribution, if restricted by the original data provider, will be handled by releasing data-construction scripts and source specifications rather than redistributing restricted raw records.

\section{Scoring Implementation Details}
\label{app:metric_details}

\paragraph{Indicator notation.}
We use $\mathbf{1}[\cdot]$ to denote the indicator function, which returns $1$ if the condition is true and $0$ otherwise.

\paragraph{Claim support for \ucr.}
Each extracted technical claim is mapped to one deterministic evidence anchor.
A claim is counted as unsupported if it contradicts the matched anchor or asserts a technical signal that is absent from the anchor.
When no verifiable claim is extracted, response-level $\mathrm{UCR}$ is set to zero by convention; such responses are not treated as reliable actions, but are penalized through the coverage term in \ndr{}.

\paragraph{Rationale polarity for \rci.}
For \rci{}, the rationale is mapped to a coarse polarity in $\{\mathrm{bullish},\mathrm{bearish},\mathrm{neutral},\mathrm{unknown}\}$.
\rci{} fires only when a directional rationale explicitly contradicts the final directional action, namely bullish reasoning with \sell{} or bearish reasoning with \buy{}.
Neutral or unknown rationales, as well as non-directional actions, are not counted as reasoning--action consistency violations.

\paragraph{Evidence strength for \eci.}
For \eci{}, evidence strength is computed from the strong-signal count $S(F)$ defined in~\cref{app:eci_signals}.
Weak evidence corresponds to $S(F)=0$, and high confidence corresponds to $c\ge4$ on the five-point confidence scale.
Thus, \eci{} fires when a model issues a high-confidence directional action under no strong verified chart signal.

\paragraph{Aggregation rationale.}
We use multiplicative aggregation because \ndr{} follows a necessary-link view of evidence-to-action reliability: reliable action requires directional coverage together with grounding, reasoning--action consistency, and evidence--confidence calibration.
The product becomes zero when any required component fails completely while remaining sensitive to graded changes across components.
Alternative aggregations exhibit different trade-offs; detailed comparisons are reported in~\cref{app:aggregation_sensitivity}.

\section{Experimental Details and Additional Results}
\label{app:experimental_details}

\subsection{Model Access and Generation Settings}
\label{app:model_settings}

All evaluations are run once under deterministic decoding whenever supported.
For open-weight and financially adapted models, we use vLLM on $8$ NVIDIA L20 GPUs with temperature $0$.
For API-based proprietary models, we set temperature to $0$ when the option is available; otherwise, we use the most deterministic setting exposed by the provider.

For reproducibility, we report the proprietary model identifiers used in our experiments:
\begin{itemize}[nosep,leftmargin=*]
    \item Gemini-3.1-Flash: \texttt{gemini-3.1-flash-lite-preview}
    \item Gemini-3.1-Pro: \texttt{gemini-3.1-pro-preview}
    \item GPT-5.5: \texttt{gpt-5.5}
    \item GPT-5.4: \texttt{gpt-5.4}
    \item GPT-5.4-mini: \texttt{gpt-5.4-mini}
    \item Claude-Sonnet-4.6: \texttt{claude-sonnet-4-6}
\end{itemize}

\begin{table}[t]
\centering
\small
\begin{tabularx}{\columnwidth}{l Y}
\toprule
\textbf{Threshold} & \textbf{Diagnostic interpretation} \\
\midrule
$\theta=0$ &
Main-paper setting.
No evidence filter is applied to directional coverage, so any directional action can contribute. \\

$\theta=1$ &
Minimal evidence threshold.
A directional action contributes only when at least one strong chart signal is present. \\

$\theta>1$ &
Stricter evidence threshold.
Coverage progressively decreases as directional actions are required to be associated with more strong chart signals. \\

$\theta\to\infty$ &
Conservative endpoint.
Coverage approaches zero for all models.
The rate of decay reflects how much directional coverage each model retains under increasingly strict evidence requirements. \\
\bottomrule
\end{tabularx}
\caption{
Operational interpretation of the evidence threshold $\theta$ in $\mathrm{NDR}_{\theta}$.
The threshold changes only the coverage aggregation rule over existing response logs.
}
\label{tab:threshold_interp}
\end{table}

\subsection{Thresholded NDR Analysis}
\label{app:threshold}

The main paper defines thresholded net decision reliability in \cref{eq:ndr} and reports \ndr{} as shorthand for $\mathrm{NDR}_{0}$.
Here, we analyze how $\mathrm{NDR}_{\theta}$ changes when the evidence threshold $\theta$ varies from $0$ to $5$.
This analysis does not require rerunning any model.
All model actions, confidence scores, extracted claims, and evidence anchors are already stored in the response logs; changing $\theta$ only changes the thresholded coverage term in \cref{eq:evidence_gate,eq:directional_coverage_theta,eq:ndr}.
The directional-conditioned chain-reliability terms remain fixed, so the decay of $\mathrm{NDR}_{\theta}$ directly reflects the loss of directional coverage under increasingly strict strong-signal requirements.

\nbf{Threshold semantics.}
\cref{tab:threshold_interp} summarizes the role of $\theta$.
The main setting $\theta=0$ applies no evidence filter to directional coverage, so any \buy{} or \sell{} action can contribute to \ndr{}.
Setting $\theta=1$ introduces a minimal evidence requirement by excluding directional actions made when no strong chart signal is present.
Larger thresholds progressively restrict coverage to directional actions associated with more strong signals.
In the limit $\theta\to\infty$, coverage vanishes for all models, while the decay rate reflects retained directional coverage under stricter evidence requirements.

\begin{figure}[!t]
\centering
\includegraphics[width=0.96\columnwidth]{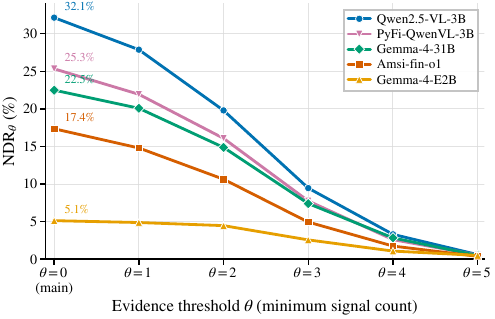}
\caption{
NDR sensitivity to evidence threshold $\theta$.
The main-paper setting is $\theta=0$, where all directional actions contribute to coverage.
Increasing $\theta$ progressively restricts coverage to directional actions associated with more strong chart signals, while the directional-conditioned chain-reliability terms remain fixed.
The leading models retain similar relative ordering under mild thresholds, whereas all models experience substantial NDR decay under stricter evidence requirements.
}
\label{fig:theta}
\end{figure}

\begin{figure}[t]
\centering
\includegraphics[width=0.9\linewidth]{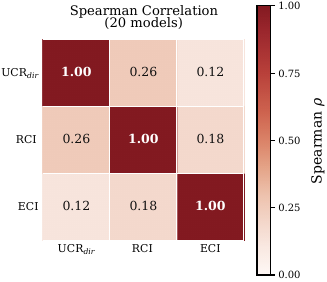}
\caption{
Local diagnostics capture non-redundant failure modes.
The heatmap reports pairwise Spearman correlations among \(\ucr_{\mathrm{dir}}\), \(\rci\), and \(\eci\) across 20 models.
All off-diagonal correlations are small (\(\rho \leq 0.26\)), indicating that factual grounding, reasoning--action consistency, and evidence--confidence calibration measure largely distinct aspects of evidence-to-action reliability.
}
\label{fig:metric_corr}
\end{figure}

\nbf{Empirical interpretation.}
As shown in \cref{fig:theta}, $\mathrm{NDR}_{\theta}$ decreases monotonically with $\theta$, as expected from the nested coverage thresholds.
The leading models retain much of their baseline NDR under mild thresholds, and their broad ordering remains stable from $\theta=0$ to $\theta=2$.
Qwen2.5-VL-3B remains the strongest model throughout this range, while PyFi-QwenVL-3B and Gemma-4-31B exhibit similar but consistently lower trajectories.
By contrast, Gemma-4-E2B starts from a low NDR at $\theta=0$ and remains low as the threshold increases, indicating that its weak reliability is not primarily caused by directional actions under zero strong signals.
Under stricter thresholds, particularly $\theta\ge3$, all models experience substantial coverage loss and their NDR values rapidly converge toward zero.
These results indicate that the main $\theta=0$ findings are not artifacts of directional actions without strong signals, while the decay under stricter thresholds characterizes each model's ability to retain reliable coverage.
Thus, $\mathrm{NDR}_{\theta}$ serves as a robustness diagnostic for \cref{eq:ndr} rather than a separate leaderboard.

\subsection{Non-Redundancy of Local Diagnostics}
\label{app:metric_correlation}

To examine whether the three local diagnostics capture complementary failure modes, 
we compute pairwise Spearman rank correlations among 
\(\ucr_{\mathrm{dir}}\), \(\rci\), and \(\eci\) across the $20$ evaluated models.
As shown in~\cref{fig:metric_corr}, all pairwise correlations are weak:
\(\rho=0.26\) for \(\ucr_{\mathrm{dir}}\)--\(\rci\),
\(\rho=0.12\) for \(\ucr_{\mathrm{dir}}\)--\(\eci\), and
\(\rho=0.18\) for \(\rci\)--\(\eci\).
These weak correlations indicate that factual grounding,
reasoning--action consistency, and evidence--confidence calibration
capture largely non-redundant aspects of evidence-to-action reliability.
A model can ground its technical claims while still contradicting its own rationale,
or remain reasoning--action consistent while being overconfident under weak evidence.
These results support the chain-level design of \bench:
a single hallucination score would conflate distinct reliability links,
whereas \ucr, \rci, and \eci provide complementary diagnostics
for identifying the dominant failure mode of each model.

\subsection{Sensitivity to Alternative Aggregations}
\label{app:aggregation_sensitivity}

We further compare the multiplicative \ndr{} with representative alternative aggregation rules on the same response logs.
Let $z_1=\mathrm{Cov}_{\mathrm{dir}}$, $z_2=1-\mathrm{UCR}_{\mathrm{dir}}$, $z_3=1-\mathrm{RCI}$, and $z_4=1-\mathrm{ECI}$.
\cref{tab:aggregation_sensitivity} summarizes the resulting behavior.

\begin{table}[!t]
\centering
\small
\begin{tabularx}{\columnwidth}{
    >{\raggedright\arraybackslash}p{0.18\columnwidth}
    >{\centering\arraybackslash}p{0.18\columnwidth}
    Y}
\toprule
\textbf{Aggregation} & \textbf{Formula} & \textbf{Key observation} \\
\midrule
Product / \ndr{} &
$\prod_{i=1}^{4} z_i$ &
Current aggregation; Gemma-4-E2B-It ranks $19/20$. \\
Arithmetic mean &
$\frac{1}{4}\sum_{i=1}^{4} z_i$ &
$\rho=0.747$ with \ndr{}; Gemma-4-E2B-It rises to rank $6/20$. \\
Minimum &
$\min_i z_i$ &
$\rho=0.889$ with \ndr{}; improvements outside the weakest component are ignored. \\
Pareto dominance &
-- &
$16/20$ models are non-dominated, and only about $2\%$ of model pairs can be ordered. \\
\bottomrule
\end{tabularx}
\caption{
Sensitivity of model comparison to alternative aggregation rules.
The multiplicative form preserves the necessary-link interpretation while retaining sensitivity to all four components.
}
\label{tab:aggregation_sensitivity}
\end{table}

The arithmetic mean permits strong components to compensate for weak directional coverage, reproducing the coverage-blind behavior that \ndr{} is designed to avoid.
The minimum preserves a strict bottleneck interpretation but discards improvements in the remaining components, while Pareto dominance provides insufficient resolution for model ranking.
These comparisons support the multiplicative form as a compact aggregation consistent with the intended evidence-to-action reliability semantics.

\subsection{Contrastive Case: Claim-Level Factuality vs. Evidence-Backed Action}
\label{app:case_ucr_ndr}

\cref{tab:case_study} provides a detailed evidence audit for a single chart-only input.
Both models receive the same input: stock 601799.SH, using the 30-day OHLCV window ending on 2023-06-30.
The ground-truth fact labels are computed deterministically from the OHLCV data.

This case illustrates why \ucr{} and \ndr{} answer different evaluation questions.
\ucr{} measures whether generated claims are unsupported, so a conservative response can appear reliable by making few verifiable claims and avoiding directional commitment.
\ndr{} instead evaluates whether the model can produce an evidence-backed action.
Gemma-4-E2B-It makes two supported claims but outputs \hold{}, whereas Qwen3.5-9B makes one unsupported volume claim among four audited claims and outputs \buy{} with confidence~4.
The latter action is coherent with the bullish evidence, is not flagged by the calibration diagnostic, and is supported by three strong signals.
Thus, \ndr{} separates conservative non-action from actionable evidence-backed reliability.

\input{tables/extra_case}

\subsection{Qualitative Case: Imperfect Claims with Reliable Action}
\label{app:case_imperfect_claims}

\begin{figure*}[t]
\centering
\includegraphics[width=\textwidth]{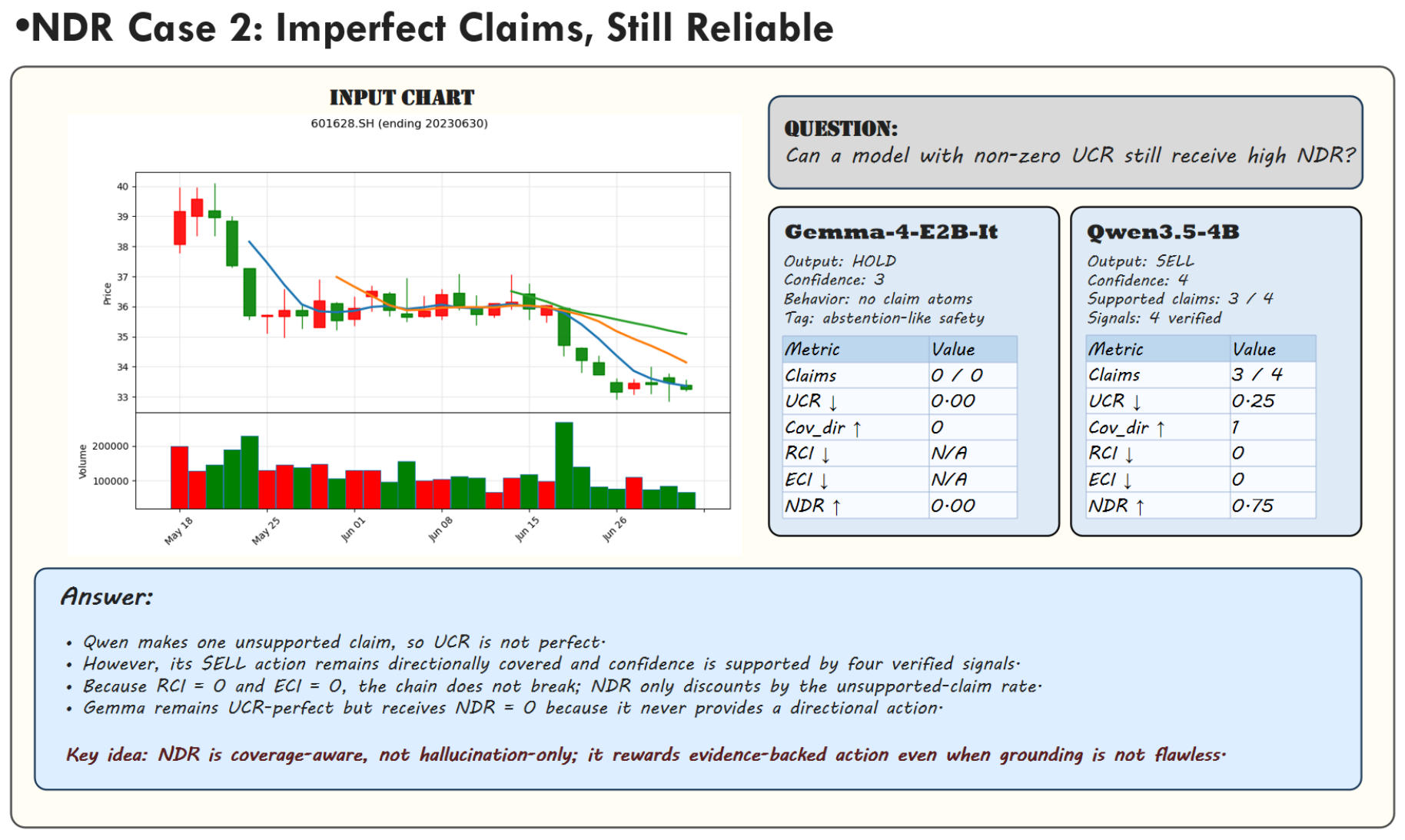}
\caption{
Imperfect claims can still yield reliable action.
Qwen3.5-4B makes one unsupported claim, resulting in non-zero \ucr{}, but its \sell{} action remains directionally covered, coherent with the bearish rationale, and supported by multiple verified evidence signals.
This example shows that \ndr{} is not a perfection-based hallucination score; it evaluates whether an actionable recommendation remains sufficiently evidence-backed across the full chain.
}
\label{fig:case_imperfect_claims}
\end{figure*}

\cref{fig:case_imperfect_claims} complements the main-text coverage-trap example by showing that \ndr{} is not a perfection-based hallucination score.
Qwen3.5-4B makes one unsupported claim, but its \sell{} recommendation remains directionally covered, coherent with the bearish rationale, and supported by multiple verified evidence anchors.
In contrast, Gemma-4-E2B-It again avoids verifiable claims and outputs \hold{}, obtaining zero observed \ucr{} but zero \ndr{} due to the absence of directional coverage.
This case illustrates that \ndr{} penalizes conservative non-action while still recognizing actionable recommendations whose evidence-to-action chain remains largely intact.

\end{document}

%% file: tables/oveall-table.tex
\begin{table*}[t]
\centering
\small
\setlength{\tabcolsep}{4pt}
\renewcommand{\arraystretch}{1.08}
\begin{tabular}{lr|rrrrr|r}
\toprule
\textbf{Model} & \textbf{Size}
& \textbf{Cov.}~$\uparrow$
& $\mathbf{UCR}_{\mathrm{all}}$~$\downarrow$
& $\mathbf{UCR}_{\mathrm{dir}}$~$\downarrow$
& $\mathbf{RCI}$~$\downarrow$
& $\mathbf{ECI}$~$\downarrow$
& $\mathbf{NDR}$~$\uparrow$ \\
\midrule
\multicolumn{8}{c}{\emph{Proprietary}} \\ \midrule
Gemini-3.1-Pro        & --  & \textbf{59.0} & 70.0 & 67.5 & 5.2 & 7.5  & \textbf{16.8} \\
Gemini-3.1-Flash      & --  & 37.9 & \underline{58.1} & 53.8 & 2.7 & 6.3  & \underline{16.0} \\
Claude-Sonnet-4.6     & --  & \underline{55.1} & 71.5 & 69.4 & 3.6 & \underline{1.9}  & \underline{16.0} \\
GPT-5.5               & --  & 25.9 & 61.2 & \underline{47.6} & \textbf{2.0} & 4.0  & 12.8 \\
GPT-5.4               & --  & 15.2 & 65.4 & 54.4 & 2.7 & 4.1  & 6.5  \\
GPT-5.4-mini          & --  & 8.3  & \textbf{55.9} & \textbf{46.5} & \underline{2.5} & \textbf{1.2}  & 4.3  \\
\midrule
\multicolumn{8}{c}{\emph{Open-Weight}} \\ \midrule
Qwen2.5-VL-3B-Instruct$^{a}$  & 3B  & \textbf{82.5} & 53.1 & 55.0 & 10.9 & 2.8 & \textbf{32.1} \\
Qwen2.5-VL-7B-Instruct$^{b}$  & 7B  & 53.9 & \underline{47.9} & \underline{48.5} & \underline{4.4} & 5.7  & \underline{25.0} \\
Gemma-4-31B-It                & 31B & 61.5 & 62.8 & 59.3 & \underline{4.4} & 6.0  & 22.5 \\
Qwen3.5-4B                    & 4B  & 52.7 & 59.1 & 54.6 & 12.3 & 3.5 & 20.3 \\
Qwen3.5-9B                    & 9B  & 64.2 & 65.0 & 63.6 & 14.5 & 5.1 & 19.0 \\
Qwen3.5-2B                    & 2B  & 54.0 & 67.1 & 61.3 & 8.4 & 2.7  & 18.6 \\
Qwen3-VL-4B-Thinking$^{c}$    & 4B  & \underline{71.9} & 75.2 & 74.1 & 6.6 & 8.8  & 15.9 \\
Qwen3.5-27B                   & 27B & 48.1 & 70.7 & 69.3 & 13.1 & \underline{1.7} & 12.6 \\
Gemma-4-E4B-It                & 4B  & 20.7 & 60.2 & 52.9 & 4.5 & \textbf{0.0}  & 9.3  \\
Gemma-4-E2B-It                & 2B  & 6.4  & \textbf{39.9} & \textbf{14.5} & \textbf{1.6} & 4.8  & 5.1  \\
\midrule
\multicolumn{8}{c}{\emph{Financial Fine-Tuned}} \\ \midrule
PyFi-QwenVL-3B-COT-47K$^{a}$  & 3B  & \underline{62.4} & \textbf{40.5} & \textbf{41.8} & 26.8 & \textbf{4.8} & \textbf{25.3} \\
PyFi-QwenVL-7B-COT-47K$^{b}$  & 7B  & 49.5 & \underline{50.7} & 54.1 & \underline{10.6} & \underline{12.5} & \underline{17.8} \\
Amsi-fin-o1$^{c}$              & 4B  & \textbf{81.3} & 69.2 & 73.7 & \textbf{5.3} & 14.2 & 17.4 \\
FinLLaVA                       & 7B  & 36.9 & \underline{50.7} & \underline{47.2} & 15.9 & 15.1 & 13.9 \\
\bottomrule
\end{tabular}%
\caption{
Main trimodal leaderboard on \bench.
Reported metrics aggregate performance over the three input modalities; NDR is computed from the aggregated components rather than by averaging modality-specific NDR values.
Cov. denotes directional coverage, and NDR denotes coverage-aware net decision reliability.
Within each group, \textbf{bold} and \underline{underlined} entries mark the best and second-best values.
Superscripts indicate the base--fine-tuned pairs: $^{a}$Qwen2.5-VL-3B, $^{b}$Qwen2.5-VL-7B, and $^{c}$Qwen3-VL-4B-Thinking.
}
\label{tab:leaderboard}
\end{table*}

%% file: tables/full-facts.tex
\begin{table*}[!t]
\centering
\scriptsize
\setlength{\tabcolsep}{3pt}
\renewcommand{\arraystretch}{1.12}
\begin{tabularx}{\textwidth}{c l l l Y Y l}
\toprule
\textbf{\#} & \textbf{Fact Key} & \textbf{Dimension} & \textbf{Inputs} & \textbf{Computation Rule} & \textbf{Output Labels} & \textbf{ECI Signal} \\
\midrule
1 & MA alignment
& MA stacking
& Close
& Compare MA5, MA10, and MA20 on the last day.
& \texttt{bullish}, \texttt{bearish}, \texttt{mixed}
& No \\

2 & MA cross
& Short-term crossover
& Close
& Detect golden/death cross of MA5 and MA10 within the last 5 days.
& \texttt{golden\_cross}, \texttt{death\_cross}, \texttt{no\_cross}
& Yes \\

3 & Volume change
& Volume anomaly
& Volume
& Compare current volume with the 20-day mean; heavy if $>1.5\times$, light if $<0.7\times$.
& \texttt{heavy\_volume}, \texttt{light\_volume}, \texttt{normal\_volume}
& Yes \\

4 & Price breakout
& Range breakout
& Close, High, Low
& Compare the evaluation-day close with the high and low
over the preceding 20 trading days, excluding the
evaluation day.
& \texttt{breakout\_high}, \texttt{breakout\_low}, \texttt{within\_range}
& Yes \\

5 & Trend direction
& Price trend
& Close
& Fit a linear trend to 20-day close prices; use $0.15\%$/day as the trend-label threshold.
& \texttt{uptrend}, \texttt{downtrend}, \texttt{sideways}
& Yes$^\dagger$ \\

6 & Volatility
& Return variability
& Close
& Compute annualized standard deviation of recent returns.
& \texttt{high\_volatility}, \texttt{medium\_volatility}, \texttt{low\_volatility}
& No \\

7 & Recent 5-day change
& Short-window return
& Close
& Compute 5-day cumulative price change; significant if $|\Delta|>3\%$.
& \texttt{significant\_rise}, \texttt{significant\_drop}, \texttt{flat}
& No \\

8 & Vol--price divergence
& Price-volume relation
& Close, Volume
& Compare 10-day price change with recent volume change.
& \texttt{bullish\_divergence}, \texttt{bearish\_divergence}, \texttt{volume\_price\_sync\_up}, \texttt{volume\_price\_sync\_down}, \texttt{no\_divergence}
& Yes \\

9 & Price position
& Range position
& Close, High, Low
& Compute the close percentile in the 20-day high--low range; high if $\geq0.80$, low if $\leq0.20$.
& \texttt{high\_zone}, \texttt{mid\_zone}, \texttt{low\_zone}
& Yes \\

10 & Short-term momentum
& Momentum acceleration
& Close
& Compare recent daily change with short-window average change.
& \texttt{accelerating\_up}, \texttt{decelerating\_up}, \texttt{flat\_momentum}, \texttt{decelerating\_down}, \texttt{accelerating\_down}
& Yes \\

\bottomrule
\end{tabularx}
\caption{
Full specification of the ten deterministic OHLCV-derived evidence anchors.
``Strong'' indicates whether the anchor contributes to the strong-signal count $S(F)$ used by \eci.
$^\dagger$Trend direction uses $0.15\%$/day for label assignment and a stricter $0.30\%$/day threshold for strong-signal counting.
}
\label{tab:factkeys_full}
\end{table*}

%% file: tables/extra_case.tex
\begin{table}[!t]
\centering
\scriptsize
\setlength{\tabcolsep}{3pt}
\renewcommand{\arraystretch}{1.08}
\begin{tabularx}{\columnwidth}{l Y}
\toprule
\textbf{Item} & \textbf{Paper-snapshot evidence} \\
\midrule
Stock / date & 601799.SH, 2023-06-30 \\
Input modality & Chart-only (candlestick PNG with MA5/MA10/MA20 and volume) \\

\midrule
\multicolumn{2}{l}{\textit{Deterministic OHLCV evidence}} \\
MA alignment &
Bullish: MA5 $=119.92$, MA10 $=118.30$, MA20 $=114.04$ \\

Volume ratio &
$0.98\times$ the 20-day mean (normal volume) \\

Price breakout &
Breakout above the prior 20-trading-day high:
current close $=123.60$, prior-window high $=122.48$ \\

Trend direction &
Uptrend; slope $=0.729\%$/day \\

5-day price change &
$+6.27\%$ (significant rise) \\

Price position &
99.2nd percentile within the most recent 20-trading-day
high--low range, including the evaluation day \\

Strong-signal count $S(F)$ &
3 \\

\midrule
\multicolumn{2}{l}{\textit{Case A: Gemma-4-E2B-It}} \\
Action / confidence &
\hold{} / 3 \\

Claim audit &
2 audited claims, both supported \\

UCR (response) &
$0/2=0.0\%$ \\

RCI / ECI &
N/A under directional-conditioned reporting \\

Role in $\mathrm{NDR}_0$ &
Non-directional; excluded from directional error terms
and does not increase directional coverage \\

\midrule
\multicolumn{2}{l}{\textit{Case B: Qwen3.5-9B}} \\
Action / confidence &
\buy{} / 4 \\

Claim audit &
Bullish MA alignment: supported;
breakout high: supported;
uptrend: supported;
heavy volume: unsupported
(actual ratio $0.98\times$, normal volume) \\

UCR (response) &
$1/4=25.0\%$ \\

RCI / ECI &
No inconsistency or calibration error detected \\

Coverage status &
Directional under $\theta=0$;
$S(F)=3$, so the action also satisfies the stricter
$\theta=1$ evidence floor \\

Role in $\mathrm{NDR}_0$ &
Directional action enters coverage and
directional-conditioned chain diagnostics \\

\midrule
\multicolumn{2}{l}{\textit{Aggregate decomposition from the trimodal leaderboard}} \\

Gemma $\mathrm{NDR}_0$ &
$6.4\% \times 85.5\% \times 98.4\% \times 95.2\% = 5.1\%$ \\

Qwen3.5-9B $\mathrm{NDR}_0$ &
$64.2\% \times 36.4\% \times 85.5\% \times 94.9\% = 19.0\%$ \\

\bottomrule
\end{tabularx}

\caption{
Released-artifact evidence audit for a contrastive chart-only case.
Gemma-4-E2B-It has zero response-level \ucr{} but does not issue a directional action.
Qwen3.5-9B incurs a $25\%$ response-level \ucr{} from one unsupported volume claim,
while its \buy{} action remains directionally covered, rationale-consistent, and calibrated.
}
\label{tab:case_study}
\end{table}